\documentclass{article}
\usepackage[utf8]{inputenc} % allow utf-8 input
\usepackage[T1]{fontenc}    % use 8-bit T1 fonts
\usepackage{hyperref}       % hyperlinks
\usepackage{url}            % simple URL typesetting
\usepackage{booktabs}       % professional-quality tables
\usepackage{nicefrac}       % compact symbols for 1/2, etc.
\usepackage{microtype}      % microtypography
\usepackage{fancyhdr}       % header
\usepackage{graphicx}       % graphics

\usepackage{PRIMEarxiv}
\usepackage{hyperref}
\usepackage{url}
\usepackage{cite}
\usepackage{tabularx}
\usepackage{booktabs}
\usepackage{todonotes}
\usepackage{array}
\newcolumntype{H}{>{\setbox0=\hbox\bgroup}c<{\egroup}@{}}
\usepackage{orcidlink}
\usepackage{booktabs,balance,subfigure}

\usepackage{cite}
\usepackage{amsmath,amssymb,amsfonts,amsthm, upgreek}
\usepackage{algorithmic}
\usepackage{textcomp}
\usepackage{xcolor}
\usepackage{todonotes}
\def\BibTeX{{\rm B\kern-.05em{\sc i\kern-.025em b}\kern-.08em
    T\kern-.1667em\lower.7ex\hbox{E}\kern-.125emX}}

\theoremstyle{definition}

\usepackage[ruled,vlined,linesnumbered]{algorithm2e}
\SetAlFnt{\small}
\SetAlCapFnt{\small}
\SetAlCapNameFnt{\small}
\SetKwComment{Comment}{ $\triangleright$\ }{}%
\SetCommentSty{footnotesize}
\SetKw{KwEach}{each}
\SetKw{KwIn}{Input}
\SetKw{KwOut}{Output}
\SetKw{KwSplit}{Split}
\SetKw{KwSample}{Sample}
\SetKw{KwUpdate}{Update}
\SetKw{KwGenerate}{Generate}
\SetKw{KwShuffle}{Shuffle}
\usepackage{algorithmic}
\algsetup{linenosize=\tiny}
 
\AtBeginDocument{%
  \providecommand\BibTeX{{%
    \normalfont B\kern-0.5em{\scshape i\kern-0.25em b}\kern-0.8em\TeX}}}

\title{Deep-BIAS: Detecting Structural Bias using Explainable AI
}

\author{
   Bas van Stein~\orcidlink{0000-0002-0013-7969}\\
   LIACS, Leiden University \\
   Leiden, The Netherlands\\
   \texttt{b.van.stein@liacs.leidenuniv.nl} \\
  \And
  Diederick Vermetten~\orcidlink{0000-0003-3040-7162} \\
   LIACS, Leiden University \\
   Leiden, The Netherlands\\
  \texttt{d.l.vermetten@liacs.leidenuniv.nl} \\
  \And 
  Fabio Caraffini~\orcidlink{0000-0001-9199-7368} \\
  Department of Computer Science \\
  Swansea University \\
  Swansea, UK \\
  \texttt{fabio.caraffini@swansea.ac.uk} \\
  \And 
  Anna V. Kononova~\orcidlink{0000-0002-4138-7024} \\
   LIACS, Leiden University \\
   Leiden, The Netherlands\\
  \texttt{a.kononova@liacs.leidenuniv.nl}
}

\begin{document}
\maketitle

\begin{abstract}
Evaluating the performance of heuristic optimisation algorithms is essential to determine how well they perform under various conditions. Recently, the BIAS toolbox was introduced as a behaviour benchmark to detect structural bias (SB) in search algorithms. The toolbox can be used to identify biases in existing algorithms, as well as to test for bias in newly developed algorithms. In this article, we introduce a novel and explainable deep-learning expansion of the BIAS toolbox, called Deep-BIAS. Where the original toolbox uses 39 statistical tests and a Random Forest model to predict the existence and type of SB, the Deep-BIAS method uses a trained deep-learning model to immediately detect the strength and type of SB based on the raw performance distributions. Through a series of experiments with a variety of structurally biased scenarios, we demonstrate the effectiveness of Deep-BIAS. We also present the results of using the toolbox on 336 state-of-the-art optimisation algorithms, which showed the presence of various types of structural bias, particularly towards the centre of the objective space or exhibiting discretisation behaviour. The Deep-BIAS method outperforms the BIAS toolbox both in detecting bias and for classifying the type of SB. Furthermore, explanations can be derived using XAI techniques.
\end{abstract}

%%
%% Keywords. The author(s) should pick words that accurately describe
%% the work being presented. Separate the keywords with commas.

\section{Introduction}\label{sect:intro}
As the amount of data and complexity of the optimisation problems continue to increase, the demand for effective heuristic optimisation algorithms also increases. Since no single heuristic algorithm is universally best \cite{NFLT}, it is necessary to benchmark these algorithms to understand which performs best under specific conditions. Most continuous optimisation benchmarks are performance-based, such as the Black-Box Optimisation Benchmark (BBOB)~\cite{hansen2021coco} test suite. These benchmarks provide information on the performance of an algorithm and how it compares with others in various situations. For example, one algorithm may excel at optimising separable functions while another performs well on uni-modal, highly conditioned functions. Resource-based benchmarks measure the number of resources (computation power, memory, and energy) required under certain conditions, but do not offer much insight into the behaviour of the algorithms under different circumstances. Behaviour-based benchmarks, on the other hand, provide additional information about how an algorithm behaves under different conditions, such as the movement of a population of candidate solutions in a swarm-based optimisation algorithm.

The previously proposed BIAS toolbox \cite{9828803} is such a behaviour-based benchmarking tool. It can be used to analyse whether algorithms or components of an algorithm induce structural bias (SB). SB is a type of bias inherent in iterative heuristic optimisation algorithms that affects their performance in the objective space. By detecting the presence, strength, and type of SB in a heuristic optimisation algorithm, it is possible to identify areas for improvement and understand under which conditions SB is less likely to occur. This information can be used to optimise the performance of these algorithms.

In this work, we propose an improved methodology for the BIAS toolbox for both the detection of structural bias and the classification of the type of SB. Instead of using $39$ statistical tests and their p-values, we propose using a single convolutional deep learning model to predict the presence and type of SB in the raw final-point distributions of various optimisation runs on a special test function $f_0$.
In this work, our aim is to answer the following research questions: %\ak{Can we not have RQs?} \bas{Sure, feel free to rephrase}

\begin{itemize}
    \item[RQ1] How to determine, using machine learning, whether a heuristic continuous optimisation algorithm suffers from SB?
    \item[RQ2] How to predict the type of SB suffered by a continuous heuristic optimisation algorithm?
    \item[RQ3] Can we generate visual explanations of the predicted SB type? 
    \item[RQ4] Is the proposed deep learning-based approach better than using statistical tests and under which conditions? 
\end{itemize}

To answer RQ1 and RQ2, we propose a deep learning approach on raw algorithm performance data (on a special test function) to identify which SB type (if any) is most likely to occur in a given optimiser. The complete SB benchmark deep learning test suite (Deep-BIAS) and data generators for different SB scenarios, are made open-source~\cite{anonymous_authors_2023_7498823}. To answer RQ3, we propose to use explainable artificial intelligence techniques adapted for the proposed deep learning approach to visualise the SB results. 
To answer RQ4, we evaluate both the proposed deep learning approach and the previous statistical test approach using a large set of 189 different artificially generated parameterised distributions containing 11 different scenarios of structural bias. In addition, we compare the results of both methods on a wide set of state-of-the-art optimisers.

%The paper is structured as follows: ...

\section{Structural bias}\label{sect:SB}
%https://arxiv.org/pdf/2301.01984.pdf
%In solving complex optimization problems, the location of the optimal solutions within the defined boundaries of the feasible domain is often unknown 
%it is crucial that the search process progresses towards regions with high-quality solutions based on the points sampled thus far, rather than the algorithm's inherent biases. 

In many complex optimization problems we have to deal with a complete black-box setting. This means we have no a priori information about which regions of the space contain good solution. In such a setting, the search has to start "from scratch", within the defined domain boundaries. The search should then be able to identify and progress towards promising regions with good values of objective function. Only points sampled thus far should steer algorithm's logic and operators in the subsequent steps of the search. This means that the algorithm doesn't inherently favour one region of the space over any other. In other words, in case the algorithm is to be deployed in a \textit{general situation}, it should be able to find high-quality solutions regardless of where they are located inside the feasible domain of the problem. The degree to which an algorithm exhibits such flexibility in locating optima is clearly among contributors to it success.

Unfortunately, detecting propensity of an iterative algorithm towards some parts of the domain is difficult due to interplay between the sampled landscape of the function and internal workings of the algorithm~\cite{Kononova2015}. In order to disentangle these two components, the test function $f_0$ has been defined as follows:

\begin{equation}
    f_0:[0,1]^n\to[0,1], \text{ where } \forall x, f_0(x) \sim \mathcal{U}(0,1).
\end{equation}

For this function, the optimum is located uniformly at random throughout the domain, and there can exist no region which is inherently better than another. As such, an unbiased search is expected to return a uniform distribution of final best solutions. If the distribution of these points is non-uniform, this indicates a \textit{structural bias} of the algorithm, as is not able to give equal importance to every region of the domain. 
Structural bias thus represents an algorithm's inflexibility. Because of this, we consider structural bias to be an undesired behaviour which potentially limits the algorithm's performance in a generic setting. 
%Obviously, one can imagine a beneficial situation when an algorithm is biased precisely towards a region containing the optima. 
%However, any kind of reverse engineering of bias onto a particular algorithms is currently impossible since 
The exact relation between the operators of an algorithm and its structural bias, and the way in which it might influence performance on different function landscapes, is poorly understood. However, through testing on $f_0$, SB can be identified, and thus potentially removed via a prudent choice of operators. 

% Exact mechanisms of formation of SB from cyclical interaction of algorithm's operators and extent to which algorithm's SB manifests itself when coupled with a wide range of function landscapes are currently poorly understood. However, with testing, SB can be identified and removed via a prudent choice of operators. 

\subsection{Existing methodology for measuring SB}\label{sect:}

Visual inspection of the distributions of the final solutions found on $f_0$, collected in multiple independent runs of the method under investigation and displayed component-wise \cite{bib:inselberg1985plane}, is the most intuitive approach to detecting SB. Although it provides a good understanding of how SB manifests itself in clearly biased and unbiased cases, such a procedure can be subject to personal interpretations and is time-consuming when a large number of images need to be generated and inspected (see \cite{Kononova2015, Caraffini2019,vStein2021_emergence} and repositories \cite{mendeleySBinOptAlgs,mendeley2021emergence}). Moreover, it is unable to provide reliable results in the presence of a mild SB or figure rendering artefacts, making it difficult to come to the right conclusion. 

The use of statistical testing methods removes the subjective component of SB inspection and leads to an automated decision-making process over a large data set of results, where the uniformity distributions of the final solutions obtained with multiple runs in $f_0$ are tested for uniformity. However, testing for uniformity is challenging, in particular when looking at a `practical' sample size $N$, which is the number of full runs to be performed in this case. Experiments carried out with $N$ up to $100$ and $\alpha=0.01$ using the Kolmogorov-Smirnov test \cite{kolomonorgov1933sulla}, the Cramer-Von Mises test \cite{CsorgoJ96} and Anderson-Darling \cite{anderson1952asymptotic} test in combination with the Benjamini-Hochberg correction method \cite{benjamin2010}, returned good results in some classes of algorithms, but, taken separately, these tests perform poorly when a wide variety of algorithms are investigated \cite{Kononova2015,Kononova2020PPSN}. The best results are obtained with $N=600$, which helps to detect SB more often but does not guarantee the detection of all different types of SB at any significance level \cite{vStein2021_emergence} - even higher values of $N$ are needed for smaller levels of significance and higher statistical power \cite{Kar2013}. Using multiple statistical tests to detect SB and its type reliably can be complex and laborious. From the point of view of a practitioner or an algorithm designer, these processes should be automated. 

\subsection{BIAS toolbox}\label{sec:toolbox}
BIAS \cite{bib:BIAS} is an open-source Python package, available from \cite{BIAS_code}, to benchmark SB in the continuous domain. The toolbox provides an SB detection mechanism based on the aggregation of the results of 39 statistical tests and a Random Forest model to identify the type of structural bias. It furthermore contains a data generator to sample data from a set of scenarios producing synthetic results; a component producing the parallel coordinate plots of the final best positions to display SB and those reporting the outcome of the decisions made with statistical analysis while detecting SB.

The BIAS toolbox is easy to use, is extendable and makes the SB detection process fast and simple, also over a large dataset of final best positions. It facilitates benchmarking of other statistical tests for detecting structural bias.

\begin{figure*}[!t]
    \centering
    \includegraphics[height=0.3\textwidth,trim=5mm 0mm 0mm 0mm,clip]{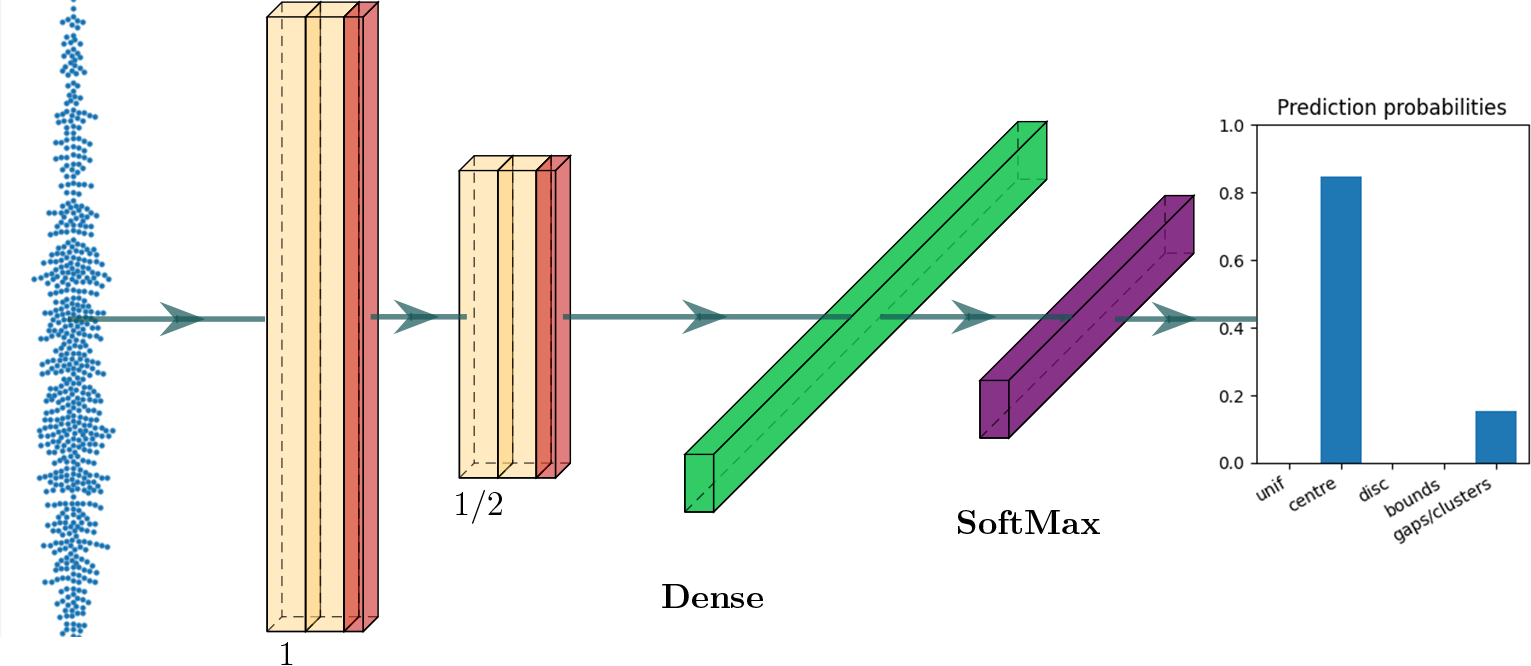}
    \caption{General one-dimensional CNN architecture with optimal hyper-parameters per sample size. The network takes as input a sorted distribution fixed sample size. Yellow layers are 1d-CNN layers, red layers are max-pooling layers, green layer is a dense layer and finally a classification head with SoftMax activation function resulting in five class probabilities per sample.\label{fig:cnn_models}
    }
\end{figure*}

\section{Methodology}\label{sect:setup}
The method proposed here extends the functionality of BIAS. It uses a deep learning-based model to predict the presence and type of structural bias. The model is trained on a large portfolio of bias scenarios and is optimised using AutoKeras. In this section, we first provide an overview of the scenarios and the generator used to train the model. Then the model training and testing procedures are explained in detail.

\subsection{Portfolio of scenarios}\label{sect:data_gen}
The proposed Deep-BIAS method is based on the parameterised SB scenarios proposed in \cite{9828803}. This portfolio of scenarios includes the most common types of structural bias as observed in previous studies, including:

\begin{itemize}
    \item bias towards the centre of the search space,
    \item bias towards the bounds of the search space,
    \item bias towards certain parts of the search space forming clusters,
    \item bias towards avoiding certain parts of the search space, creating gaps and
    \item strong discretisation.
\end{itemize}

In the original paper, there are $11$ different scenarios with different parameter settings, giving in total a set of $194$ parameterized scenarios.
After visual and analytical analysis of the parameters of these scenarios, we removed $5$ of these parameter settings, as they frequently generated distributions visually and statistically equivalent to random uniform distributions using only $600$ samples. This leaves a total of $189$ parameterized distribution generators (still $11$ scenarios), which we use to generate train and test data sets.

To train the deep learning model, we divide the $11$ scenarios into four classes, \emph{Center, Bounds, Gaps/Clusters} and \emph{Discretisation}. We argue that gaps and cluster bias are highly overlapping, as you have gaps in the search space when there are clusters and vice versa. Therefore, they are added together under one class label.
For each class $20.000$ distributions are generated, equally divided over the parameter settings and scenarios that belong to each class label. For the uniform (no bias) label, we generated $80.000$ distributions to make the prediction task balanced between bias and non-bias samples. In total, this gives a data set of $160.000$ one-dimensional distributions, of which $80\%$ is used as training and $20\%$ as final validation (using stratified sampling).

\subsection{Optimized Convolutional Neural Network}\label{sect:cnn}
In the next step, a deep one-dimensional convolutional neural network (1d CNN) is trained and optimised using the AutoKeras \cite{jin2019auto} algorithm.
Each distribution is first ordered in ascending order of the values and then used as input for the 1d CNN. AutoKeras is set to run with an evaluation budget of $100$ trials and without image augmentation and pre-processing (this would have no meaning in our situation since we are not dealing with images). Each network is trained for $50$ epochs and uses a small randomly selected validation set (from the original training data) to compare different models with each other. The best neural network instance is stored and used in this work.
The general architecture of the models is visualised in Figure \ref{fig:cnn_models}. There are two blocks, each containing two convolutional layers followed by a max pooling layer. The second block is half the size of the first. Followed by a dense layer and a SoftMax classification head. In total, there are four models, one for each sample size of 30, 50, 100 and 600. 
Training and optimisation of the networks took roughly one day per sample size on an NVIDIA T4 GPU.

\subsection{Predicting SB and SB type}\label{sec:rf_model}
The optimised CNN models can now be used to directly predict any distribution for the presence and type of structural bias.
First, we examine the models using the final validation set left out.

\begin{figure*}[!htb]
    \centering
    \includegraphics[height=0.27\textwidth,trim=8mm 17mm 25mm 20mm,clip]{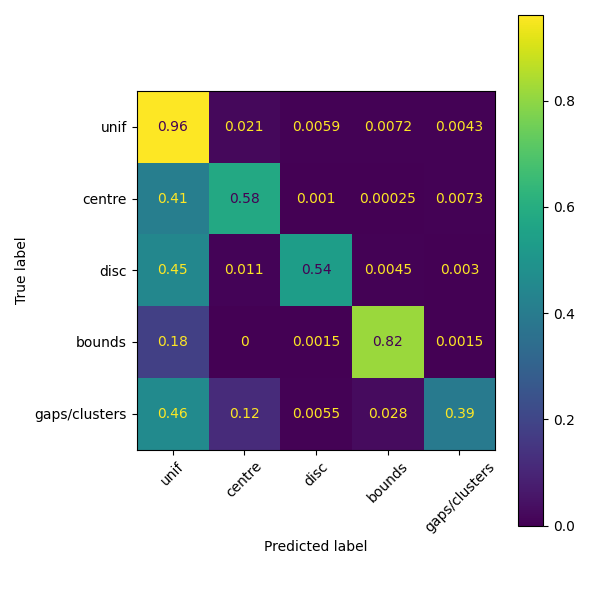}
    \includegraphics[height=0.27\textwidth,trim=35mm 17mm 25mm 20mm,clip]{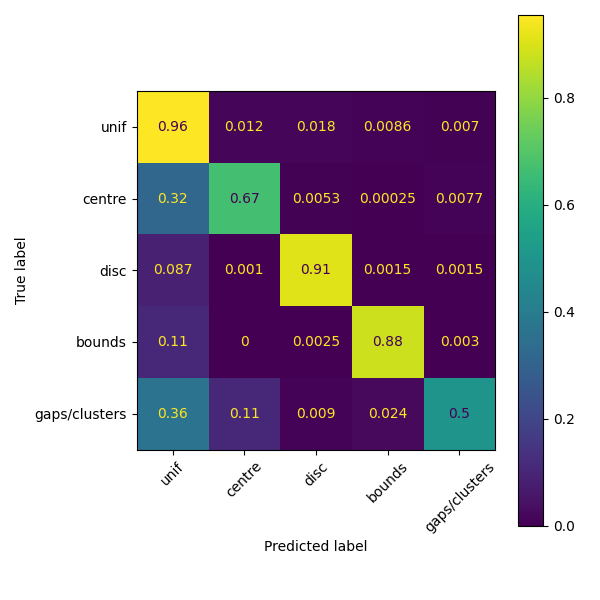}
    \includegraphics[height=0.27\textwidth,trim=35mm 17mm 25mm 20mm,clip]{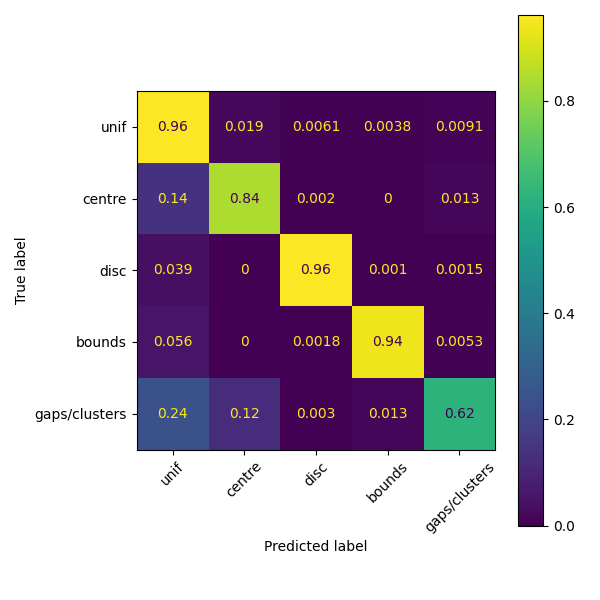}
    \includegraphics[height=0.27\textwidth,trim=35mm 17mm 25mm 20mm,clip]{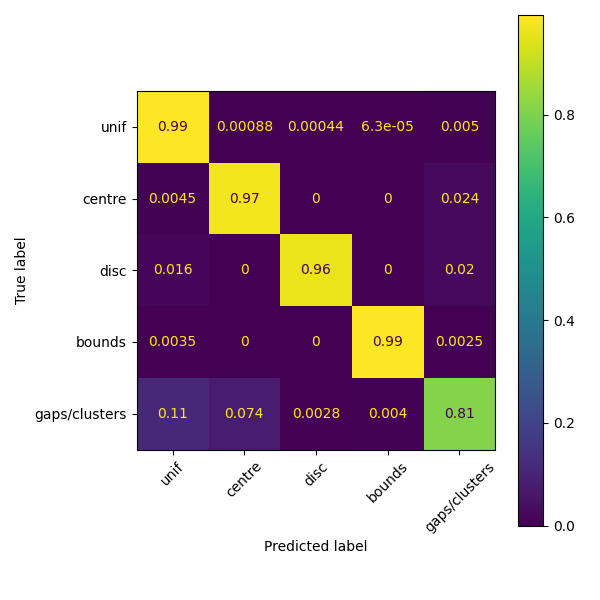}
    \includegraphics[height=0.27\textwidth,trim=130mm 17mm 0mm 0mm,clip]{confusion/opt_cnn_model-600-confusion.png}
    \caption{Confusion matrices for the models with sample sizes from left to right: 30, 50, 100 and 600. The horizontal axis denotes the Predicted labels and the vertical axis the True labels. Trained on a data set of 128.000 one-dimensional distributions and validated on a set of 32.000 distributions. The 600 samples model has a macro F1 score of 0.95.
    \label{fig:confusion}
    }
\end{figure*}

In Figure \ref{fig:confusion}, the classification results for each sample size are shown. It can be seen that the classification results obviously improve with the number of samples, which is also the case for the statistical approach proposed in \cite{9828803}. Interestingly, even the $50$ samples network is doing a good job of predicting uniform distributions as uniform. Most misclassifications occur when the model predicts the distribution to be uniform, but this is not the case.

\begin{figure*}[!ht]
    \centering
    \includegraphics[width=0.49\textwidth,trim=4mm 4mm 4mm 4mm,clip]{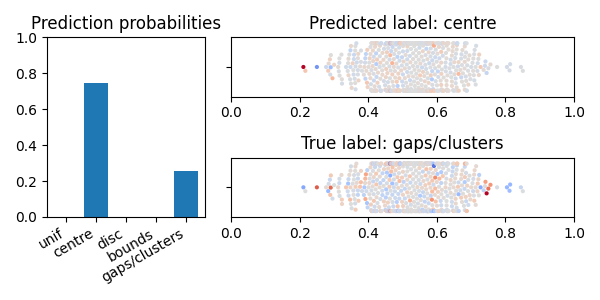}
    \includegraphics[width=0.49\textwidth,trim=4mm 4mm 4mm 4mm,clip]{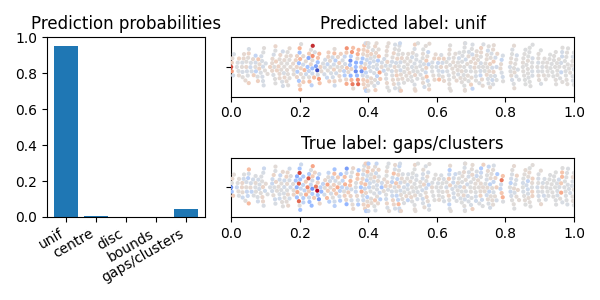}
    \includegraphics[width=0.49\textwidth,trim=4mm 4mm 4mm 4mm,clip]{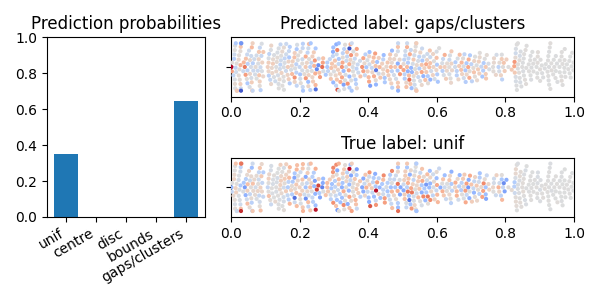}
    \includegraphics[width=0.49\textwidth,trim=4mm 4mm 4mm 4mm,clip]{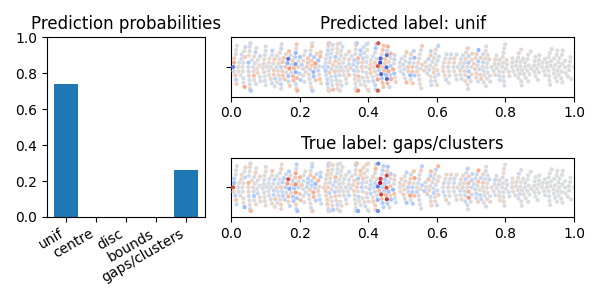}
    \caption{Examples of (wrongly) classified samples due to overlap in classes. Clusters can be located by chance on the bounds or in the center of the space, small gaps can occur randomly in uniform sampling by chance. Colours indicate Shapley values where dark red indicates a positive contribution towards the class label (either predicted (up) or ground truth (bellow)), and dark blue indicates a negative contribution. Sample points with similar values are stacked on top each other.
    \label{fig:misclassifications}
    }
\end{figure*}

To further analyse the misclassifications of each of the models, we used the explainable artificial intelligence (XAI) method, SHAP \cite{shapley_approx}, which approximates Shapley values \cite{Shapley1953} based on a background sample set from training data. Using the Shapley values for each point in the distribution, it is possible to highlight regions of interest for a particular prediction.
In Figure \ref{fig:misclassifications}, four examples of predictions are shown that do not match the ground truth. These examples come from the $600$ sample size model. It is important to note that there is a high level of randomness in many scenario generators. It can therefore occur that clusters are very much overlapping, creating a uniformly distributed sample, or that clusters can be located in the centre or at the bounds, which deceives the classifier.
In the examples shown, it is also visible that some of the generators can generate such subtle deviations (tiny gaps in a uniform distribution) that they are hard to classify correctly, as these can also occur randomly in a random uniform distribution by chance for $600$ samples.

Next, the Deep-BIAS method is compared to the original BIAS toolbox.
The comparison is done by transforming the problem into a binary problem to detect SB. Since the original BIAS toolbox is also validated in this way on a different number of dimensions (not just one-dimensional distributions), the same experiment is repeated here.
The following dimensions are evaluated: 1, 10, 20 and 30.
For each of these dimensionalities ($d$), a test set of $1890$ biased and $1890$ unbiased $d$-dimensional distributions are generated.
These distributions are then predicted by both the original BIAS toolbox and the newly proposed Deep-BIAS method. For the proposed method, the class probabilities are averaged over all dimensions to give a final prediction per $d$-dimensional distribution. We define a configuration as biased if at least $10\%$ of its dimensions are classified as non-uniform. This threshold is chosen to remain consistent with the original BIAS toolbox~\cite{9828803}.

\begin{figure*}
    \centering
    \includegraphics[width=0.98\textwidth,trim=10mm 10mm 10mm 5mm,clip]{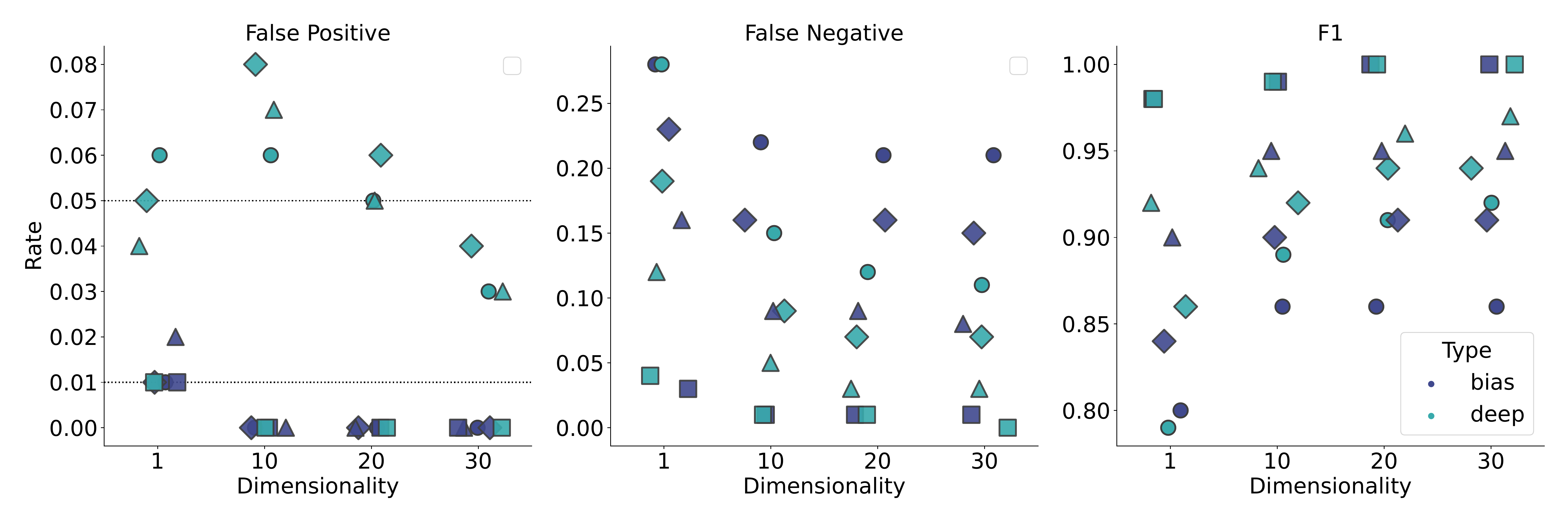}
    \caption{Comparison (with $\alpha=0.01$) of the original BIAS toolbox (blue) and the Deep-BIAS (teal) in terms of false positives (left), false negatives (middle) and F1-score (right). On all figures, markers identify the used sample size: $\circ$, $\diamondsuit$, $\triangle$ and $\square$ are 30, 50, 100 and 600, respectively.} %\ak{please make font larger}
    \label{fig:fpr_fnr_f1}
\end{figure*}

The results of this experiment are shown in Figure~\ref{fig:fpr_fnr_f1}, where false positives, false negatives, and F1 scores are compared for different dimensions and sample sizes. We can see that while Deep-BIAS has a slightly higher rate of false positives, this is compensated by a significantly lower false negative rate. In general, the F1 score for Deep-BIAS is slightly higher, indicating that it slightly outperforms the statistical approach.  

% @Diederick: Insert figure with false positives, false negatives per d. \dv{Figure~\ref{fig:fpr_fnr_f1}}

\section{Benchmarking structural bias of real algorithmic data}\label{sect:real-world}

This section benchmarks a large set of heuristic optimisation algorithms by applying the BIAS toolbox, answering RQ3.

\subsection{Data collection setup}
We use data from a heterogeneous pool of heuristics executed on $f_0$ at dimensionality $n=30$ for a maximum of $10000\cdot n$ fitness functional calls. In total, we consider $336$ optimisation heuristics that fall into the following categories (all use $N=100$):
\begin{itemize}
    \item Variants of Differential Evolution ($195$ configurations), 
    \item Compact optimisation algorithms ($81$ configurations), 
    \item Single-solution algorithms ($60$ configurations), 
    % \item Variants of Genetic Algorithms ($96$ configurations). 
\end{itemize}
For the sake of clarity and reproducibility, the exact composition and setup of these algorithmic configurations are fully described in a dedicated document available from \cite{BIAS_code}.

\subsection{Results}

\begin{figure*}
    \centering
    \subfigure[DE configurations]{\includegraphics[height=0.70\textwidth,trim=10mm 9mm 30mm 9mm,clip]{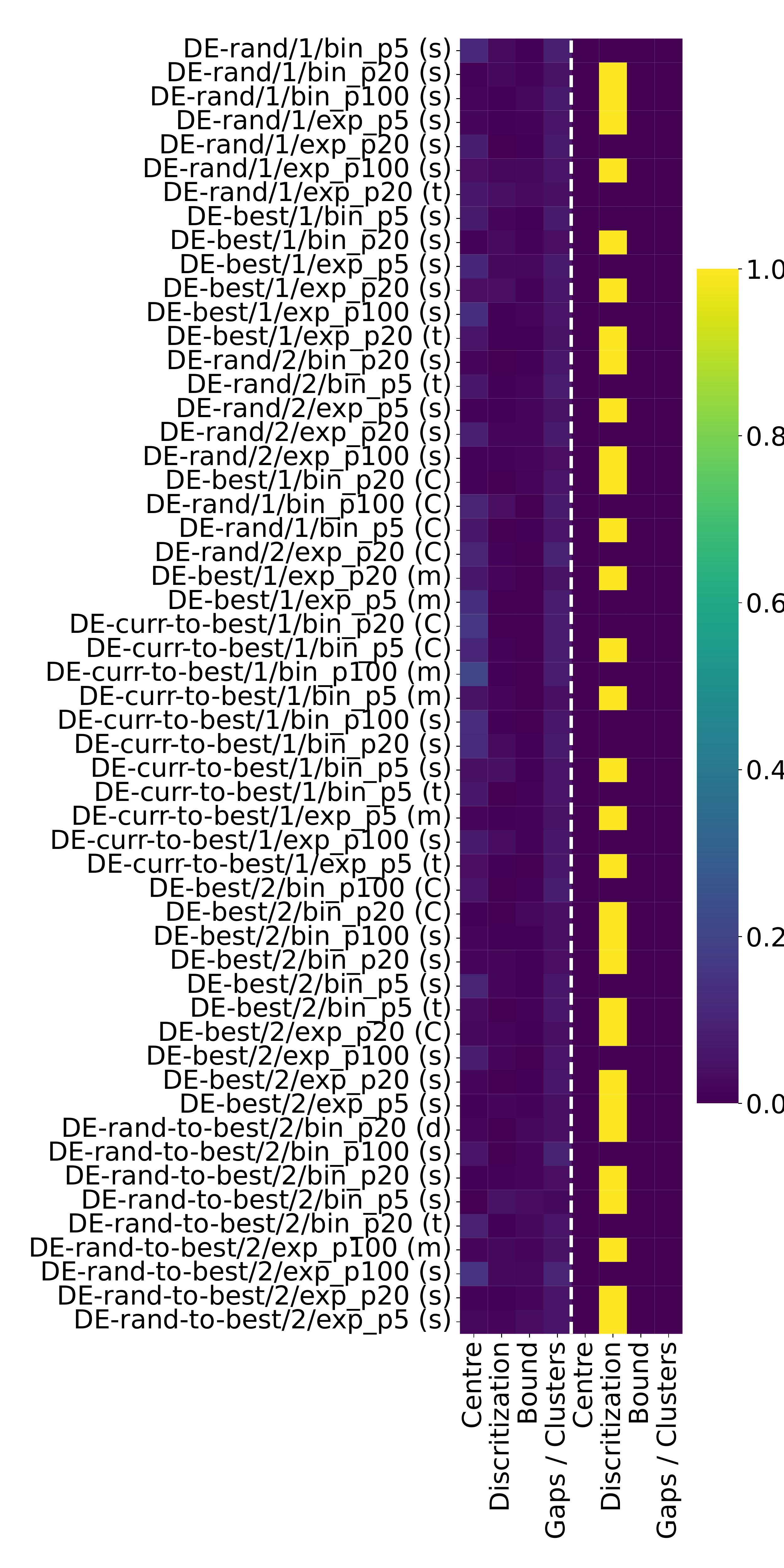}
    %\caption{Predictions of types of non-uniformity made by the deep (left) and original (right) bias methods on the set of DE configuration where the two methods give a different biased / non-biased outcome. For the original method, the probabilities are determined by the random forest model when the configuration is considered biased and set to 0 otherwise. For deep, the predictions are the average of the per-dimension predictions.}
    \label{fig:classes_deep_rf_DE}}
    \subfigure[Single solution algorithms]{\includegraphics[height=0.70\textwidth,trim=10mm 9mm 1mm 9mm,clip]{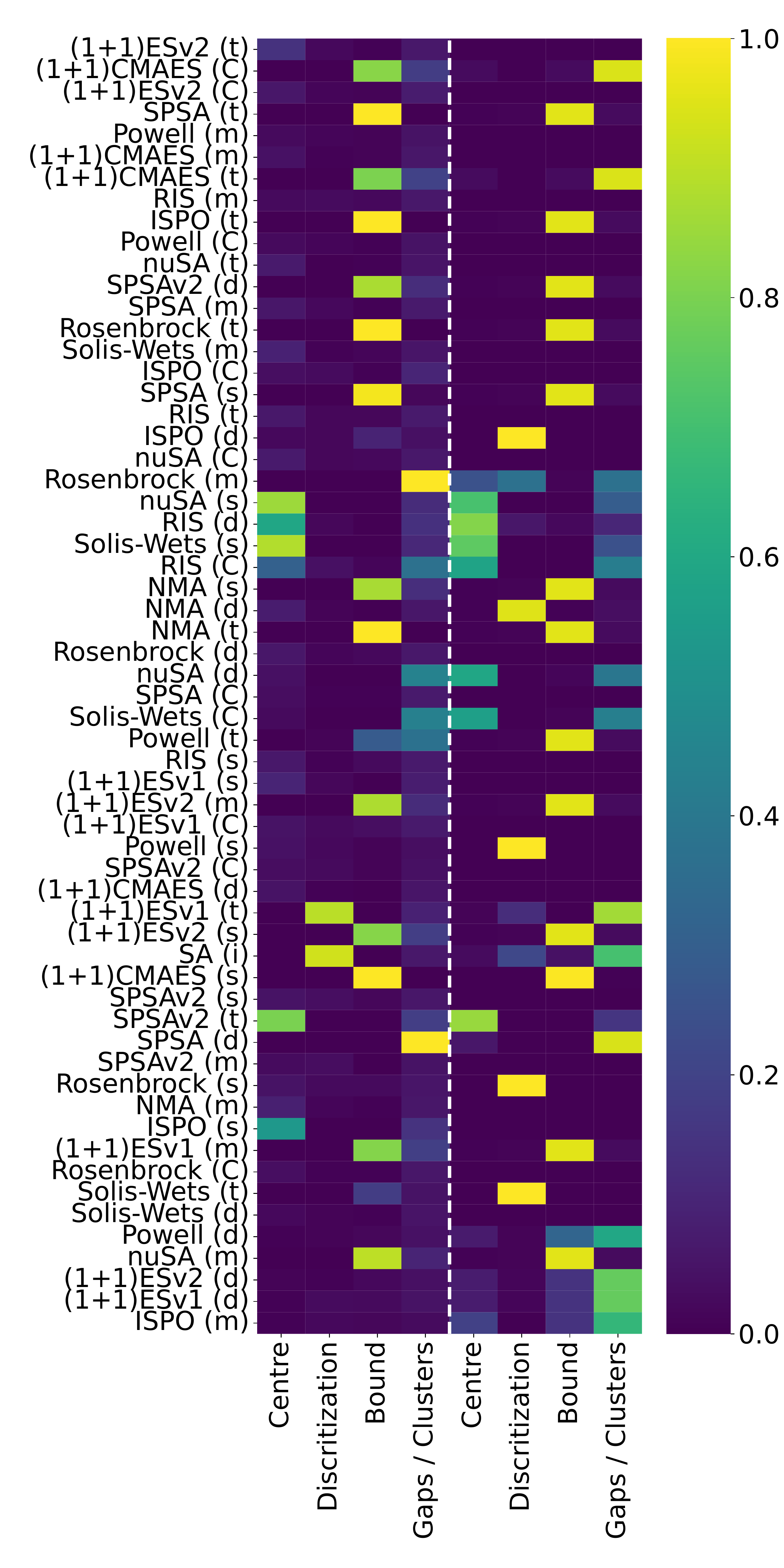}
    %\caption{Predictions of types of non-uniformity made by the deep (left) and original (right) bias methods on the set of single solution algorithms. For the original method, the probabilities are determined by the random forest model when the configuration is considered biased and set to 0 otherwise. For deep, the predictions are the average of the per-dimension predictions.}
    \label{fig:classes_deep_rf_algs}}
    \caption{Predictions of types of non-uniformity made by the Deep-BIAS (left of white dash line) and original (right of white dash line) bias methods on the sets of DE configurations (left) and single solution algorithms (right) where the two methods give a different biased / non-biased outcome. For the original method, the probabilities are determined by the random forest model when the configuration is considered biased and set to 0 otherwise. For Deep-BIAS, the predictions are the average of the per-dimension predictions. The colour bar on the right applies to both sub-figures.}
\end{figure*}

First, we compare the decisions made by the Deep-BIAS method with those of the original toolbox on the set of single-solution algorithms. This is done by showing the class probabilities of both methods for each algorithm in Figure~\ref{fig:classes_deep_rf_algs}. From this figure, we see that, in most cases, both methods give the same biased/non-biased outcome. However, the type of bias detected varies for a large part of the algorithms. This often occurs when the random forest (RF) model of the original toolbox predicts `clusters'. This might be due in part to the differences in training data for the two models, combined with the fact that for some cluster settings, the clusters might be located near the bounds, making the distinction between these two classes somewhat fuzzy. 

In addition to the comparison of the single-solution algorithms, we can also investigate the impact on different versions of Differential Evolution to gain insight into the relationship between the algorithm configuration and the types of bias detected. In this setting, we highlight only the algorithm for which the two models make different biased/non-biased decisions and show the resulting class probabilities in Figure~\ref{fig:classes_deep_rf_DE}. It is clear that there are two distinct cases here: for most models where the RF decides that the configuration is biased, it predicts `discretisation'. This matches the fact that these configurations all use the `saturate' boundary correction, which places any infeasible solution back on the boundary. As such, if an algorithm generates some fraction of infeasible solutions, there is a high probability that at least some of the coordinates are placed exactly on the boundary, resulting in several equal values for some coordinates. This is highly unlikely for a uniform distribution, so statistical tests easily detect this discrepancy.  

For Figures~\ref{fig:classes_deep_rf_algs} and \ref{fig:classes_deep_rf_DE}, it is also important to note that for the RF model, the class probabilities sum up to 1 by design, which is not the case for the deep model, as the uniform class still gets some of the probability mass. Thus the outcome of Deep-BIAS gives indirectly a measure of the strength of the bias. 

\begin{figure*}
    \centering %height=0.374\textwidth,trim=9mm 10mm 9.4mm 9mm,clip
    \subfigure[DE-best/1/bin\_p20 (s)]{\includegraphics[height=0.674\textwidth,trim=9mm 16mm 221.4mm 9mm,clip]{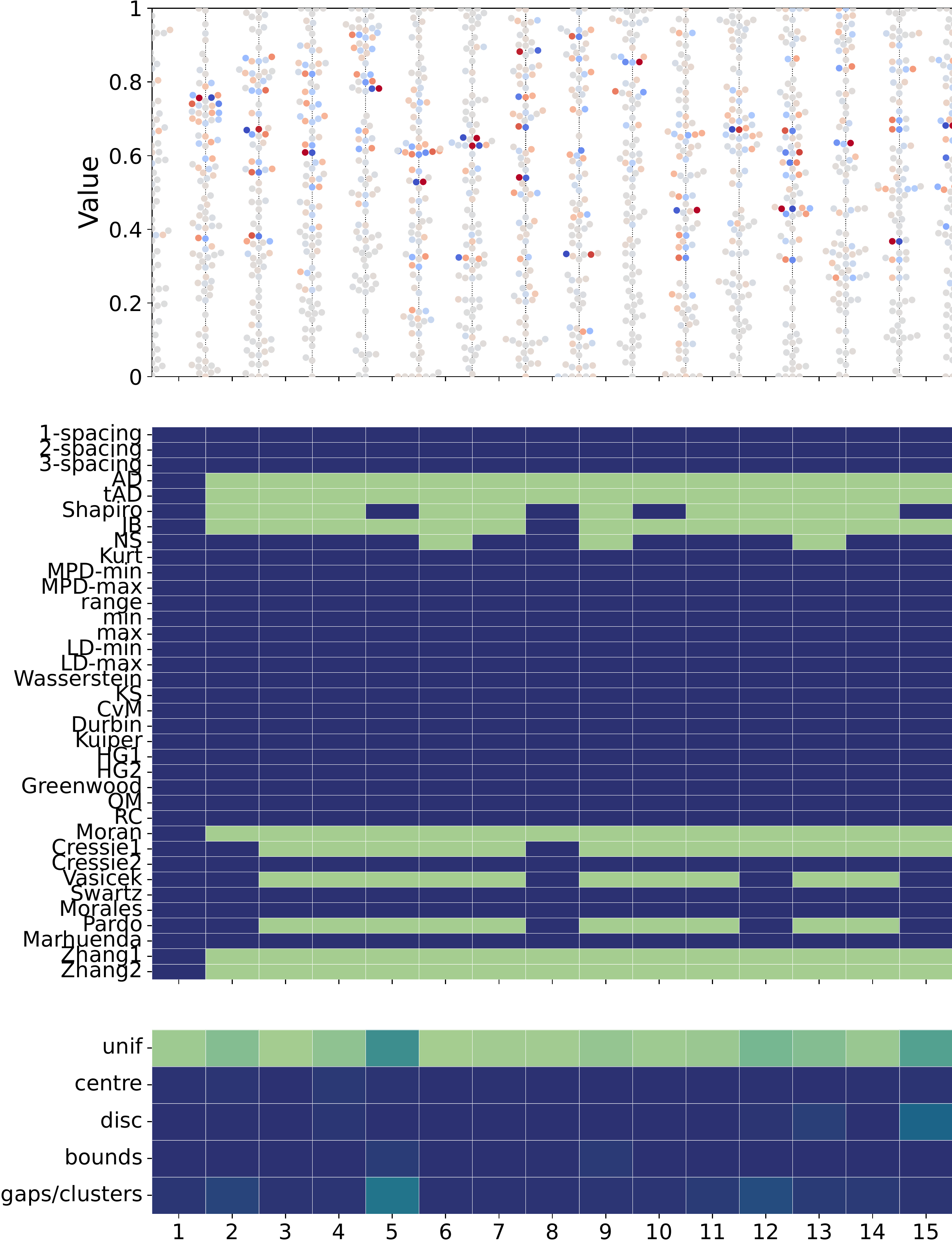}
    %\caption{Overview of the positions (top), statistical test results (middle) and deep predictions (bottom) for the 'DE-best/1/bin\_p20 (s)' algorithm. Colours in the top part are based on the SHAP values as discussed in Section~\ref{sec:rf_model}. \ak{please make font larger}}
    \label{fig:expl_197}}
    \subfigure[Solis-Wets (C)]{\includegraphics[height=0.674\textwidth,trim=49mm 16mm 221.4mm 9mm,clip]{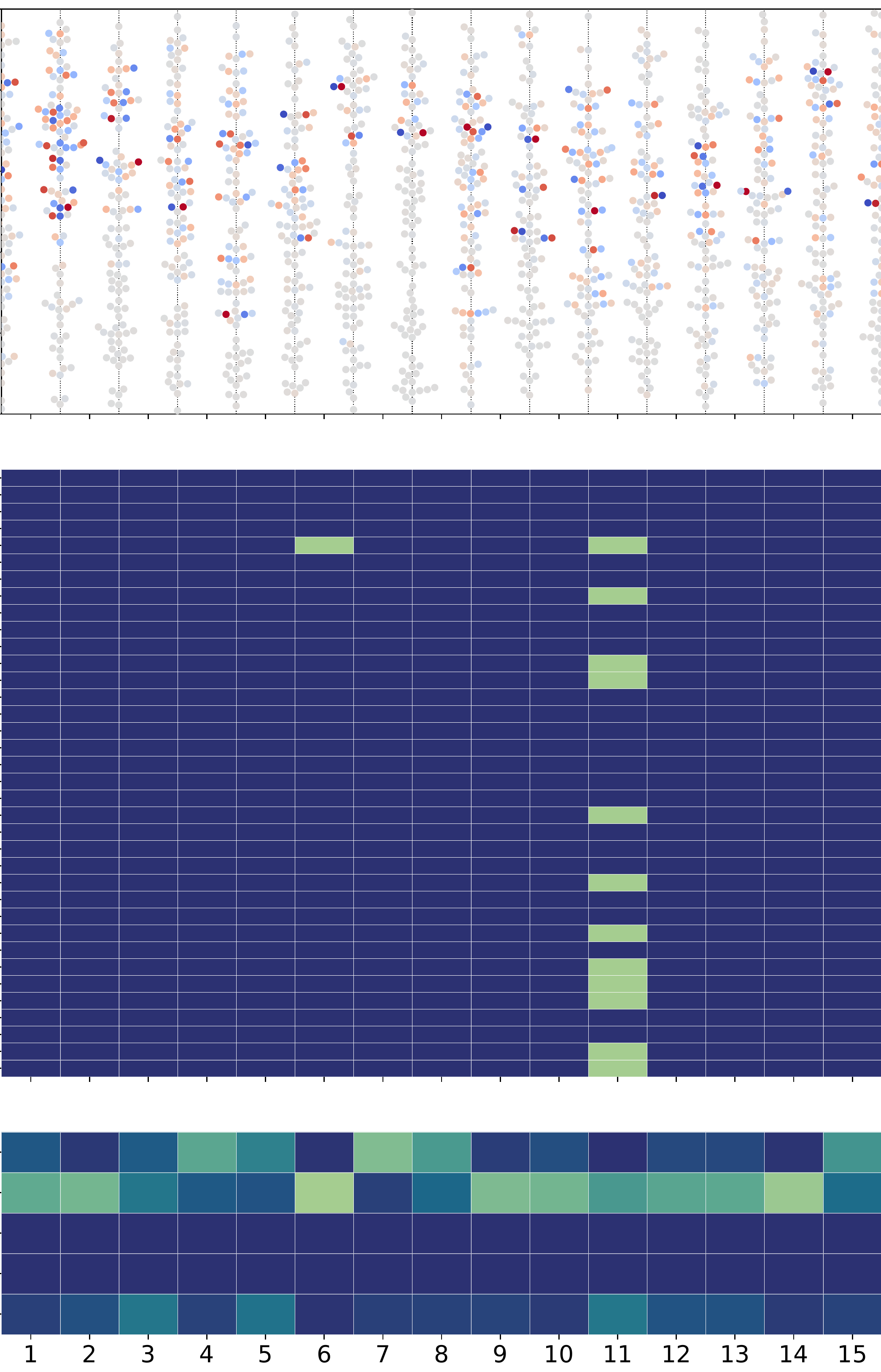}
    %\caption{Overview of the positions (top), statistical test results (middle) and deep predictions (bottom) for the 'Solis-Wets (C)' algorithm. Colours in the top part are based on the SHAP values as discussed in Section~\ref{sec:rf_model}. \ak{please make font larger}\dv{this is the largest the testnames can be without overlapping (there is already some overlap now)}}
    \label{fig:expl_468}}
    \caption{Overview of the positions (top), statistical test results (middle) and deep predictions (bottom) for two algorithms (dimensions 1 to 15 out of 30 are shown). Colours in the top part are based on the SHAP values as discussed in the caption of Figure~\ref{fig:misclassifications}. Legend on the left applies to both subfigures.}\label{fig:expl_both}
\end{figure*}

%\dv{Someone please check these next two paragraphs, I'm getting a bit lost on what I can repeat where and how it fits in the overall story} \bas{looks good}
To gain more insight into the scenarios where the Deep-BIAS model differs from the original toolbox, we can zoom in on two configurations where their predictions disagree. Figure~\ref{fig:expl_both} shows two scenarios which highlight the  differences in behaviour between the two methods for detecting bias. In Figure~\ref{fig:expl_197}, we show a DE-variant for which the statistical tests clearly detect non-uniformity, while the Deep-BIAS model finds no strong evidence for structural bias. When looking at the distribution of the points, we notice that many dimensions have multiple points which lie exactly at the border. This is a natural effect of the `saturate' boundary correction method, since this places points generated outside the domain on the closest boundary. Since the probability of a uniform distribution generating a point exactly on the border is $0$, the cases where multiple points are on the bound are seemingly easy to detect by a wide variety of tests. The Deep-BIAS model however, does not have this context, so it is less sensitive to the saturation of the boundaries.

In Figure~\ref{fig:expl_468}, we show the reverse case: the statistical tests are not rejected, but the Deep-BIAS model consistently classifies the points as non-uniformly distributed. Interestingly, the distribution visually seem to have slightly more mass in the centre of the space. However, the deviations are very small, and as such the tests are not confident enough to be able to reject the null-hypothesis after the necessary p-value corrections are applied. Since this case does not have enough test-rejections, we might argue that it can be considered as non-biased. However, since the deep model is not limited to binary decisions, we can instead make use of the resulting probabilities to highlight this setting as having a very low level of bias. This shows how both methods can be used together to make more informative decisions on the presence, strength and type of structural bias. 
% \hl{Diederick}

\section{Reproducibility statement}
We provide an open-source documented implementation of our Deep-BIAS package on Github \cite{anonymous_authors_2023_7498823}.
Experiments, additional figures, and documentation can be found there. Pre-trained models are available on the same repository under the `models' directory.
All models are trained using a single T4 GPU and the computations are carbon neutral ($CO_2$-free) by using solar power.

%Some findings from running the tests: 
%\begin{itemize}
%    \item There were a few issues with the correction method in the code of BIAS. I did the correction in the wrong axis, resulting in too heavy across-dimension correction, but no per-dimension corrections. This has been resolved, and should in theory mean we don't need the across-dimension limit anymore, although I think it might actually be rather useful still. We could apply the same limit for the deep setting, as that seems to remove some edge cases where one dimension is just barely seemed as biased (for both methods this occurs sometimes: 45 cases for original, 84 for deep)
%    \item The Moran test is a bit strange, for a few (10) of the cases where nothing else rejects, it rejects on basically all dimensions. 
%    \item Depending on how I count, the deep detects bias in 166 cases vs 154 for the original (with Moran removed, across-dimension limit kept)
%    \item There are quite a few (94) cases where there is a disagreement between the two methods. Interestingly, it seems deep fails to find bias when saturate is used, but does trigger for the other corrections which BIAS did not consider problematic.
%\end{itemize}

\section{Conclusion}\label{sect:conclusion}
In light of the analysis performed in this study, we conclude that the use of deep learning is a viable option to detect SB with satisfactory results. With only $50$ samples, we can correctly detect most uniformly distributed points, and performances increase with larger sample sizes. We find that the optimal network architecture for detecting SB is not as complex as those often designed in other deep learning applications in the literature, and yet it behaves similarly to the original BIAS toolbox, which is based on statistical tests, and performs very well in terms of the F1 score evaluation metric.

Compared to BIAS, the proposed Deep-BIAS alternative displays some interesting features and advantageous behaviours.  
\begin{itemize}
    \item It outperforms the statistical test-based approach in classifying the type of SB.
    \item It gives a better measure of strength of the SB by using the class probabilities.
    \item It provides additional insights by using XAI, where regions of interest in distributions can be further analysed to understand the mechanism behind the classifications.
\end{itemize}
A disadvantage of using a neural model is that the resulting SB detection system is less generalisable than that obtained with a statistical test approach, where the network might miss SB types that are not considered in the training data. However, the training data set prepared for Deep-BIAS appears to be adequate, and the network fails to find bias mainly when the saturate constraint handling method is used. This is not a problem for BIAS. For these reasons, the overarching conclusion of this study envisages the joint use of the 2 systems for optimal SB detection. We recommend using BIAS as the primary method for binary bias/non-bias classification and Deep-BIAS to inspect the type of SB and determine its strength.

%%
%% The acknowledgments section is defined using the "acks" environment
%% (and NOT an unnumbered section). This ensures the proper
%% identification of the section in the article metadata, and the
%% consistent spelling of the heading.

%\begin{acks}
%To Robert, for the bagels and explaining CMYK and color spaces.
%\end{acks}

%%
%% The next two lines define the bibliography style to be used, and
%% the bibliography file.
\bibliographystyle{unsrt}
\bibliography{main}

%%
%% If your work has an appendix, this is the place to put it.
\appendix

\end{document}